%% file: mquat-mds.tex
\documentclass[dvipsnames, sigconf]{acmart}

\usepackage[utf8]{inputenc}
\usepackage{amsmath}
\usepackage[english]{babel}
\usepackage{microtype}
\usepackage{graphicx}
\usepackage[obeyFinal]{todonotes}
\usepackage{booktabs}
\usepackage{csquotes}
\usepackage{subcaption}
\usepackage{pgfplotstable}
\usepackage{url}
\usepackage{multicol}
\usepackage{amssymb}
\usepackage{multirow}
\usepackage{pifont}

\newcommand{\cmark}{{\color{OliveGreen!70!black}\ding{51}}}%
\newcommand{\xmark}{{\color{red!80!black}\ding{55}}}%

\input{lst}

\setcopyright{rightsretained}

\acmISBN{}
\renewcommand\footnotetextcopyrightpermission[1]{}
\makeatletter
\def\blfootnote{\xdef\@thefnmark{}\@footnotetext}
\makeatother

\begin{document}
\title{Parameter Tuning for Self-optimizing Software at Scale}
%\titlenote{Produces the permission block, and copyright information}
%\subtitle{Extended Abstract}

%\author{Dimitrii Pukhkaiev, Sebastian G\"otz, Serhii Kashuba, Johannes Mey, Rene Sch\"one, Karsten Wendt}
%\affiliation{%
  %\institution{Software Technology Group, Technische Universit\"at Dresden, Germany}
%}
%\email{(firstname.lastname)@tu-dresden.de, sebastian.goetz@acm.org}

\author{Dmytro Pukhkaiev}
\affiliation{%
  \institution{Software Technology Group, Technische Universit\"at Dresden, Germany}
}
\email{dmytro.pukhkaiev@tu-dresden.de}

\author{Uwe Aßmann}
\affiliation{%
	\institution{Software Technology Group, Technische Universit\"at Dresden, Germany}
}
\email{uwe.assmann@tu-dresden.de}

\renewcommand{\shortauthors}{D. Pukhkaiev et al.}

% Taking notes
\newcommand{\dimitriic}[1] {\todo[inline,color=orange]{DP: #1} }

% Hightlighting
\newcommand{\nttype}[1]{\textbf{#1}}

\input{abstract}




\maketitle

\input{intro}
\input{rqs}
\input{background}
\input{concept}

\input{eval}

\input{threats}
\input{conclusion}
\input{acknowledgements}

\bibliographystyle{ACM-Reference-Format}
\bibliography{mquat-mds} 

%\cleardoublepage
%\pagebreak
%\input{backup}

\end{document}

%% file: lst.tex
\usepackage{listings}
\usepackage{xcolor}

\lstdefinestyle{common-style}{%
	basicstyle=\ttfamily\scriptsize,       % the size of the fonts that are used for the code
	numbers=left,                   % where to put the line-numbers
	numberstyle=\ttfamily\scriptsize,      % the size of the fonts that are used for the line-numbers
	stepnumber=1,                   % the step between two line-numbers. If it is 1 each line will be numbered
	numbersep=5pt,                  % how far the line-numbers are from the code
	showspaces=false,               % show spaces adding particular underscores
	showstringspaces=false,         % underline spaces within strings
	showtabs=false,                 % show tabs within strings adding particular underscores
	frame=single,                   % adds a frame around the code
	tabsize=2,                      % sets default tabsize to 2 spaces
	captionpos=b,                   % sets the caption-position to bottom
	breaklines=true,                % sets automatic line breaking
	breakatwhitespace=false,        % sets if automatic breaks should only happen at whitespace
	keywordstyle={\color{blue}\textbf},         % keywords are blue, (and blue)
	commentstyle={\color{OliveGreen}},          % comments
	literate={\$}{{{\$}}}1 {<=}{{$\leq$}}1 {>=}{{$\geq$}}1,
	escapechar=\&,
	basewidth=0.5em
}

\lstdefinelanguage{JRAG}[]{java}{%
morekeywords={abstract,public,private,boolean,aspect,null,syn,inh,coll,eq,with,int,contributes,new,return,for,if,else,this,to,true,false}
, morecomment=[l]{//}, morecomment=[s]{/*}{*/}, }

\lstdefinelanguage{AST}{%
morekeywords={abstract, ::=}
, morecomment=[l]{//}, morecomment=[s]{/*}{*/},
}

\lstdefinelanguage{jastadd-mquat}[]{java}{%
 morekeywords={component, using, property, contract, requires, resource, of, type, requiring, providing, mode},
 morecomment=[l]{//},
 morecomment=[s]{/*}{*/},
 style=common-style,
}

%% file: abstract.tex
\begin{abstract}
	
	Efficiency of self-optimizing systems is heavily dependent on their optimization strategies, e.g., choosing exact or approximate solver. A choice of such a strategy, in turn, is influenced by numerous factors, such as re-optimization time, size of the problem, optimality constraints, etc. Exact solvers are domain-independent and can guarantee optimality but suffer from scaling, while approximate solvers offer a ``good-enough'' solution in exchange for a lack of generality and parameter-dependence. In this paper we discuss the trade-offs between exact and approximate optimizers for solving a quality-based software selection and hardware mapping problem from the scalability perspective. We show that even a simple heuristic can compete with thoroughly developed exact solvers under condition of an effective parameter tuning. Moreover, we discuss robustness of the obtained algorithm's configuration. Last but not least, we present a software product line for parameter tuning, which comprise the main features of this process and can serve as a platform for further research in the area of parameter tuning.
	
\end{abstract}

%% file: intro.tex
\section{Introduction}
\label{sec:intro}

Combinatorial\blfootnote{To appear in Workshop on Model Selection and Parameter Tuning in Recommender Systems (MoST-Rec'19)} optimization problems are repeatedly occurring in all spheres of life. Navigation and route planning for autonomic vehicles~\cite{WAN2016548}, planning and scheduling of production tasks~\cite{pinedo2012scheduling} and sport events~\cite{KENDALL20101}, recommender systems~\cite{JJL17} and many more. Naturally, the decades-long attention to these topic brought numerous approaches and algorithms for solving one or another problem from this list.

As a running example for this paper we use a problem of quality-based software selection and hardware mapping~\cite{ttc18}. It is an important problem occurring in the field of self-optimizing software~\cite{GKPPA-14}. In short, it requires a satisfaction of functional and nonfunctional requirements of the users by serving their requests at a certain point in time. A request is specified in a form of a contract that requires a variant of a software component providing a certain functionality. The software components themselves can require other software components and a hardware instance to be deployed on. The solution is an optimal~(w.r.t. a quality measure, e.g., overall energy consumption) mapping of software variants onto the available hardware, whilst satisfying all contracts.

This problem was used as a case of the Transformation Tool Contest 2018\footnote{\url{https://www.transformation-tool-contest.eu/2018/}}~(TTC), where several promising solutions were presented. The results of TTC'18 are very expressive as the best two approaches are diametrically different by their nature: a heuristic-based ant-colony optimization approach~\cite{HD18} and an exact integer linear programming-based~(ILP) approach~\cite{GMSA-18}. The result continues a decades-long competition between exact and approximate solvers~\cite{Kuenne1972}. The general claim is that ``good enough'' heuristics have much better scaling factor than exact solvers. However, the recent advancements in computational resources made exact solvers very attractive for tackling optimization problems. For example, in~\cite{GMSA-18} the authors showed the superiority of an ILP-based solution to the heuristic~\cite{HD18} in terms of scalability; thus, arguing the utilization of heuristics.

In this paper we aim to back up the usage of approximate solvers for combinatorial optimization problems. The main problem of such approaches lies in costly and often not obvious parameter optimization~(tuning)~\cite{Birattari2009}. Choosing an appropriate parameter configuration can significantly improve performance of an algorithm. Therefore, we present \texttt{BRISE 2}, a software product line~(SPL) for parameter tuning, which effectively reduces the effort spent on optimizing the algorithm's parameters. Its idea was inspired by a successful application of a similar approach in the energy benchmarking area~\cite{PG18}.%, which was able to reduce the effort on finding an energy-optimal configuration up to 80\%.

To show the power of the automatic parameter tuning we introduce a simple heuristic based on a simulated annealing algorithm~\cite{khachaturyan1979statistical} that randomly varies software and hardware components. We pick an initial configuration according to the established guidance, which is already a manual parameter tuning. BRISE 2 was able to further improve the solution's quality by 11\%~(for larger problem sizes). We also show how and under which conditions a combination of parameter tuning with such a heuristic can be beneficial for solving the quality-based software selection and hardware mapping in comparison to the exact ILP-based solution~\cite{GMSA-18}.

%Optimal vs heuristic 

%Problem:

%\begin{itemize}
%	\item Problem of EAT, scalability and performance issues
%	\item Formulation of the optimization problem
%	\item Utilization of ILP, untuned MH as a solver
%\end{itemize}

%Goal:

%Scalable solving of the aforementioned problem.

%Solution:

%BRISE + MH

%% file: rqs.tex
\section{Research Questions}
The goal of this paper is to motivate the utilization of heuristics for the problem of quality-based software selection and hardware mapping. The research objective is to determine the conditions under which a utilization of a heuristic-based solution can be beneficial. To reach the research objective we need to answer the following research questions.

\begin{itemize}
	\item[\textbf{RQ1.}] What is the effect of parameter tuning on the solution quality of the approximate algorithm?
	\item[\textbf{RQ2.}] Is it possible for a tuned heuristic to provide a comparable solution to the ILP-based solver?
	\item[\textbf{RQ3.}] What is the scalability of the heuristic in comparison to ILP-based solver?
	%\item RQ4. How costly is the process of parameter tuning itself in comparison to a utilization of the ILP-based solver or an untuned heuristic?
\end{itemize}

%% file: background.tex
\section{Quality-based software selection and hardware mapping and its Solvers}
\label{sec:background}

\subsection{Problem definition}

Let us discuss the problem domain of the paper: the problem of quality-based software selection and hardware mapping~\cite{ttc18}. This optimization problem is twofold. It presumes finding an optimal serving of incoming requests by the means of mapping the corresponding software components~(SWC) onto available resources, hardware components~(HWC) w.r.t. energy-efficiency.  

Assume a scenario where the user requires a video service with a specified non-functional properties~(NFP) such as, e.g., a minimal resolution~(1440~p) and a frame rate~(60~fps). The required functionality can be provided by a core SWC: \textsc{VideoPlayer}. Each SWC can, in turn, have a set of different \textit{implementations} such as, e.g, VLC~(Video LAN Client) and QT~(Quicktime). Moreover, a SWC can require other SWCs, e.g., a \textsc{VideoPlayer} SWC requires a \textsc{Decoder} and \textsc{DataProvider}. A HWC is a hierarchical structure, where a component \textit{server} comprises several sub-components: \textit{CPUs}, \textit{RAM}, \textit{disk}, \textit{network}. 

The optimal solution consists of a selection of the best SWC implementations and their best mapping to the available HWCs for a given set of requests.

\subsection{Exact and approximate solvers}

The classic solution to the aforementioned problem is: Multi-Quality Auto-Tuning (MQuAT)~\cite{GKPPA-14}. It is an approach for model-driven, component-based development of self-adaptive systems. The key idea is a utilization of model transformation from a problem specification to a reasoning logic. The problem specification is described with a domain specific language, while the reasoning logic is represented in a form of an ILP.

MQuAT is an example of a two-phased optimization approach with two distinct parts: domain- and solver-specific. In this case the problem modeling is done by a domain expert in domain-specific part, whilst after the transformation the described model is domain independent and can be solved by a standard constraint solver.

Heuristic-based solvers are the algorithms that are situated solely in the domain-specific part. They are enriched with domain knowledge and therefore are able to effectively find a valid solution. However, they do not guarantee optimality and in the long run are often losing to the exact solvers in terms of solution's quality. %Figure~\ref{taf} depicts the difference between the approaches.    

%\begin{figure}[t]
%	\centering
%	\missingfigure{MQuAT + ILP, MH in TAF-like way}
%
%	\caption{TAF}
%	\label{taf}
%\end{figure}

\subsection{Metaheuristic-based solver}

In this paper we present a simple heuristic based on the simulated annealing meta-heuristic that solves the stated problem. It is developed with help of OptaPlanner\footnote{\url{https://www.optaplanner.org/}}, a lightweight, embeddable planning engine. The choice of an underlying meta-heuristic is very important for the resulting quality of the solver. However, in this paper we want to show that even a straightforward heuristic can deliver solutions comparable and even outperform thoroughly developed ILP-based solvers. Thus, we leave a question of heuristic selection to specialized works~\cite{Antosiewicz2013, Wortmann2017, BOUSSAID201382}. First, let us discuss the main concepts of the approach.

\begin{figure*}[t]
	\centering
	\includegraphics[width=1\linewidth]{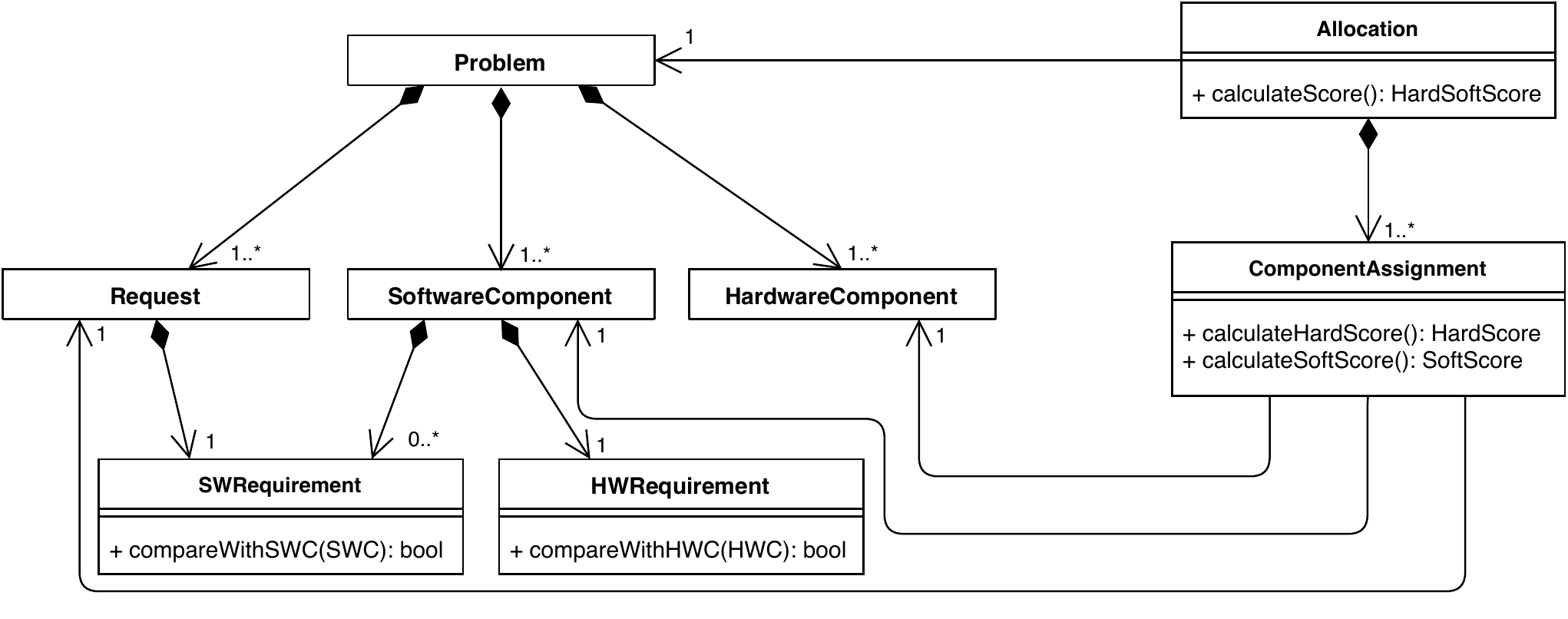}
	\caption{Class diagram of our metaheurisic-based solver}
	\label{fig:MH-CD}
\end{figure*}

Figure~\ref{fig:MH-CD} depicts a class diagram of the heuristic. \textsc{ComponentAssignment} is a placeholder for interchangeable variables, which in context of our case study are \textsc{HWCs} and \textsc{SWCs}. Each \textsc{ComponentAssignment} contains exactly one \textsc{SWC-HWC} pair and the \textsc{Request} it corresponds to. A list of \textsc{ComponentAssignments} forms an \texttt{Allocation}, the optimization of which is our final goal. Note, there are several unintuitive design decisions that were made in order to make the problem model compatible with OptaPlanner. For example, from the problem description we know that different implementations of the same SWC can have an arbitrary number of requested components. Therefore, the overall number of \textsc{ComponentAssignments} can vary depending on the current choice of the solution. OptaPlanner, however, needs to create all planning entities in advance to manipulate them during the search. Hence, we created the maximum number of \textsc{ComponentAssignments} and enable or disable them depending on component requirements of a current solution.

Each iteration of any local search~(LS) is based on two main notions: \emph{move} and \emph{score calculation}. Move is a change of a planning variable that provides a new solution, i.e., explores the search space. Score calculation is necessary to assess the quality of the obtained solution and, thus, helps to decide which move to take. 

\begin{figure}[h]
	\centering
	\includegraphics[width=1\linewidth]{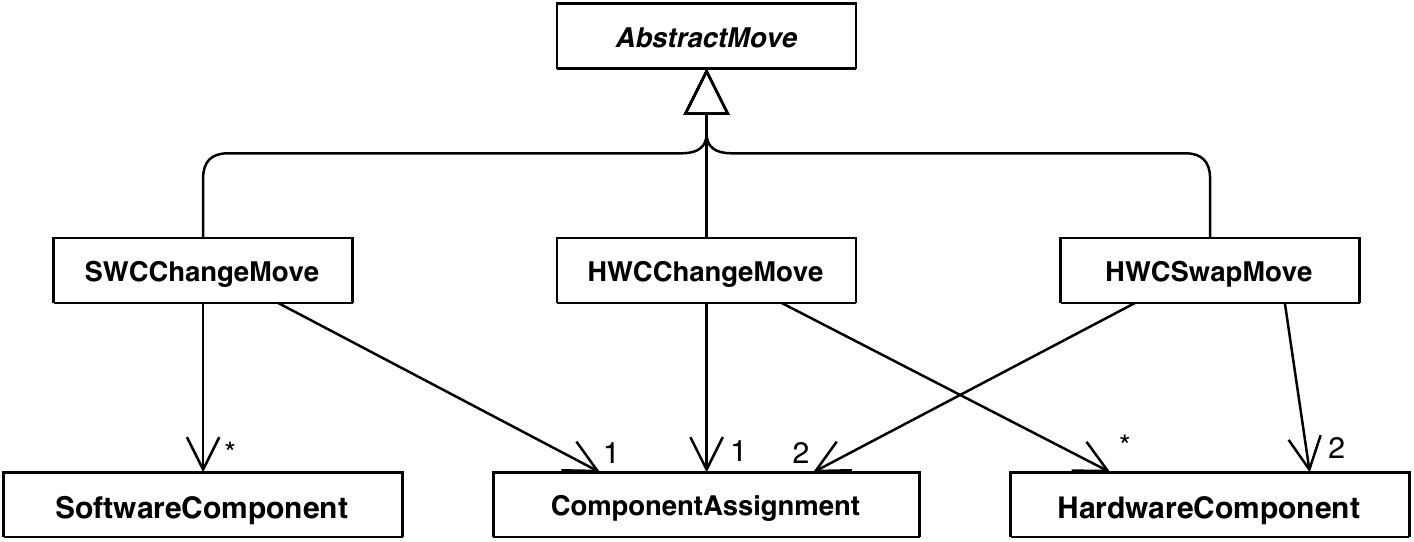}
	\caption{Class diagram of the possible moves}
	\label{fig:move-cd}
\end{figure}

There are three types of moves in our MH: \textsc{HWC-change move}, \textsc{HWC-swap move} and \textsc{SWC-change move}. Change move means a random substitution of an assigned HWC or SWC by a different one, while swap move switches an HWC of one \textsc{ComponentAssignment} by the content of the another one and vice versa. An \textsc{SWC-change move} is a bit more complicated and deserves additional details. After changing a random \textsc{SWC} at a random level of hierarchy it causes a modification of requirements for the dependent \textsc{ComponentAssignments}, removal of the obsolete assignments and addition of the new ones. Note, that the addition of the new assignments implies not a creation of new \textsc{SWC-HWC} pair, but unblocking an existent \textsc{ComponentAssignment}, thus, allowing to utilize them at the current stage of the LS. Figure~\ref{fig:move-cd} shows a class diagram of possible moves.

After the move is performed, the score calculation takes place. We iterate through all \textsc{ComponentAssignments} and check whether the underlying \textsc{SWCs} and \textsc{HWCs} are complying with the stated requirements. We utilize a so-called ``hardsoft'' score\footnote{\url{https://docs.optaplanner.org/7.21.0.Final/optaplanner-docs/html_single/index.html\#hardSoftScore}} in our MH. It allows to decouple the objective function~(energy consumption) from other requirements while comparing two solutions. If any constraint, such as missing or unsatisfying qualities of an \textsc{SWC} or \textsc{HWC}, is violated, the hard score is increasing. The soft score summarizes the energy consumption of all \textsc{HWCs} executing picked \textsc{SWCs}. The optimal score looks as follows: 
\begin{equation}
0 \, \texttt{hard} \, / \, min(energy\_consumption) \, \texttt{soft}.
\end{equation}

Note, that the score $0 \,  \texttt{hard} \, / \, 1000 \, \texttt{soft}$ is better than \\ $1 \,  \texttt{hard} \, / \, 10 \, \texttt{soft}$ as the hard score has a higher priority than the soft. By using such a scoring strategy we can ensure the knowledge whether the obtained solution is valid~(serves all requests).   

The employed simulated annealing algorithm, which provides a basis for our heuristic, has a strong dependence from its parameters. Picking the right parameter configuration can make the algorithm highly effective, while a poor choice can prevent search space exploration at all.

Our heuristic has the following input parameters:

\begin{itemize}

	\item \emph{subComponentUnassignedFactor}: Factor of influence for an unsatisfying HWC sub-component. The number of unsatisfying sub-components is multiplied by the chosen factor and added to the overall hard score. Possible values: [1, 2, 3, 5, 10, 100, 1000, 10000];

	\item \emph{softwareComponentUnassignedFactor}: Factor of an unsatisfying SWC.  The number of unsatisfying SWCs is multiplied by the chosen factor and added to the overall hard score. Possible values: [1, 2, 3, 5, 10, 100, 1000, 10000];

	\item \emph{hardScoreStartingTemperature}: Maximal accepted hard score degradation of a new solution in comparison to the current solution. Is relative to the hard score of the initial solution.  Possible values: [1, 2, 3, 5, 10, 20, 30, 50, 75, 100];

	\item \emph{softScoreStartingTemperature}: Maximal accepted soft score degradation of a new solution in comparison to the current solution. Possible values: [1, 2, 3, 5, 10, 20, 30, 50, 75, 100];

	\item \emph{neighborhoodSize}: Size of the neighborhood, i.e., how many neighboring solutions are being compared each iteration. Possible values: [1, 2, 5, 10, 20, 30, 40, 50].

\end{itemize}

 The total number of configurations needed to be measured to find an optimal parameter combination equals 51200 configurations for a single problem size. Assume a testing time of just 10 seconds~(in the TTC case it was 15 minutes) and a single-threaded execution the minimum required time effort equals to almost 6 days of continuous execution. In this paper we present an approach that allows to find a near-optimal configuration for the same problem size in 11 minutes, thus decreasing this effort by 99.9\%. But first we need to discuss state-of-the-art approaches of parameter tuning.

%% file: concept.tex
\section{Parameter Tuning}
\label{sec:parameter_tuning}

\subsection{Background}

Parameter tuning is a well-known problem arising in different areas, such as \emph{machine learning}~\cite{Feurer2019}, where a selection of an optimal configuration can significantly influence quality of the production phase, \emph{search-based software engineering}~\cite{AFC11}, where an effective parameter configuration can dramatically speed-up search for an optimal solution and \emph{benchmarking process}~\cite{PG18}, where it can minimize execution time, energy consumption or maximize quality of an algorithm.

Approaches that aim to reduce the number of measurements needed to identify an optimal configuration can be divided into three major groups: factorial designs~\cite{dean2015handbook, fedorov2013}, combinatorial optimization approaches~\cite{YI2015256,camilleri2014} and active learning approaches~\cite{PG18,BOHB-18,Alipourfard2017}.

\emph{Fractional factorial design} is an identification and removal of parameters that do not have much influence on the resulting quality~\cite{dean2015handbook}. It is similar to feature selection in machine learning, and even though it was originally developed for another goal~(reduction the measurements' number), fractional factorial design is applicable in this area~\cite{TANG201847}. Being able to drastically reduce the search space in case of a high parameter number, these approaches lack flexibility while dealing with remaining parameter combinations, giving no guidance on their exploration. Thus, factorial designs should be complemented with some other strategy.

\emph{General purpose combinatorial optimization algorithms} such as evolutionary algorithms or meta-heuristics are widely used in parameter tuning~\cite{HUANG2007516,YI2015256,camilleri2014}. However, due to the nature of combinatorial optimization itself, they are suitable to only a limited subset of experiments~\cite{PG18}. The optimization problem should have a huge search space, while time of identifying a resulting value for a single configuration should be as small as possible. Such a scenario allows to compare a vast number of the neighbors to find a solution for the next step. Being applied to a long-running target algorithm, these approaches are unforgivably wasteful. 

Active learning approaches such as~\cite{PG18, BOHB-18, Alipourfard2017, Dalibard2017, Golovin2017} iteratively sample the search space and construct the model which is used to predict the next configuration to be measured, thus, keeping the number of measured configurations at each point of time as small as possible. These approaches have received the highest attraction during the recent years. We refer the interested reader to the respective papers for details and describe a generalized approach in this paper.

\subsection{Generalization and Parameter Tuning SPL}

A typical active learning approach looks as follows. After getting a search space from the user, the algorithm starts with \emph{sampling of configurations}, this sampling can be random~\cite{Golovin2017, PG18, BOHB-18, Alipourfard2017} or adhere to some specific strategy~\cite{PG18}. After each measured configuration the \emph{model} of the search space based on the measured configurations is created. It can be Bayesian~\cite{Dalibard2017, BOHB-18, Golovin2017}, direct-acyclic-graph-based~\cite{Dalibard2017}, linear-regression-based~\cite{PG18}, etc.  The process of building the model starts with a very small data set. Thus, it can be very inaccurate and one needs a metric to \emph{validate} it~\cite{PG18}. It is based on a comparison of possible results for the task, specified by a user and of predicted results by the model. After the model is successfully validated, sampling is dropped and the model leads the search into the promising areas of the search space. Each newly found \emph{solution candidate} should also be \emph{validated} to determine the algorithm's termination~\cite{Golovin2017, PG18, Alipourfard2017}, otherwise, a user-defined timeout is another popular stopping criteria~\cite{BOHB-18}.

\begin{figure*}[t]
	\centering
	\includegraphics[width=1\linewidth]{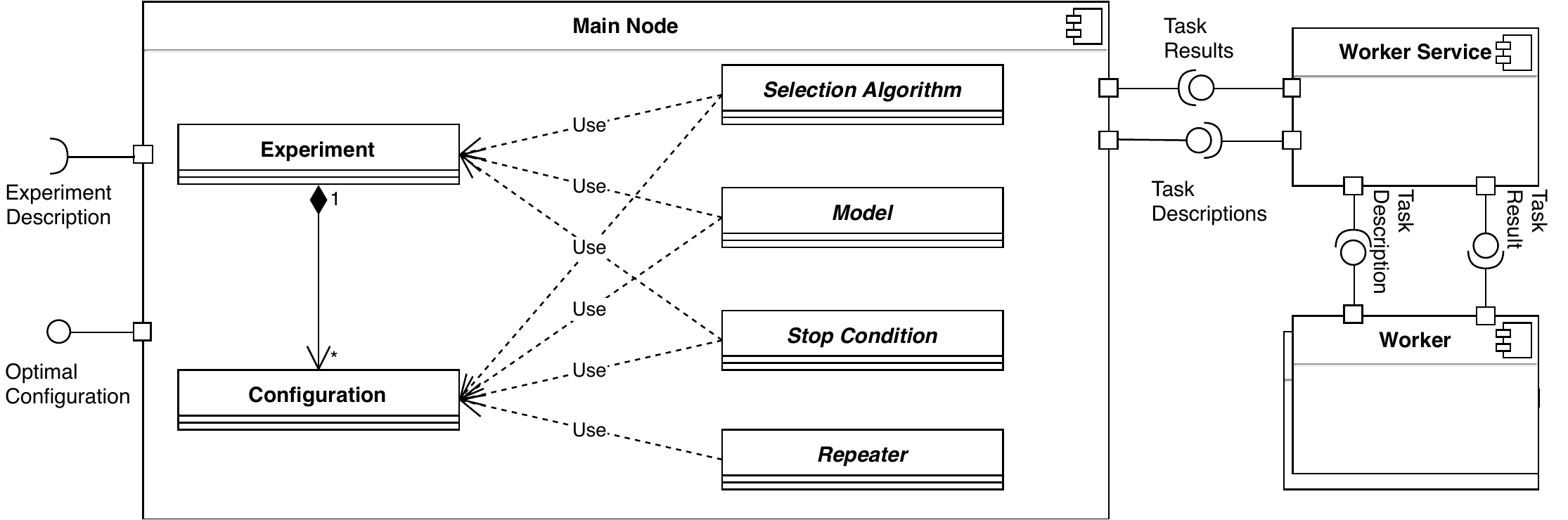}
	\caption[]{High-level architecture of the SPL}
	\label{fig:brise-cd}
\end{figure*}

Unlike its predecessor BRISE 1~\cite{PG18}, BRISE 2 uses this generalized flow as a basis, has multiple variants for each previously described feature, is easily extensible; and thus, is a full-fledged software product line for parameter tuning. Figure~\ref{fig:brise-cd} depicts a high level architecture of the SPL. 

BRISE 2 adheres to the client-server architecture and consists of three main components. \emph{Main Node} is a core component of the framework, it is responsible for the whole flow of the application except the measurement process. \emph{Worker} is a lightweight component, containing only the logic of the algorithm to be tuned (or its wrapper), which should be specified by the user. It gets a parameter combination to be tested as an input and outputs the quality metric to be later evaluated in the \emph{Main Node}. The communication between the \emph{Main Node} and \emph{Workers} is performed via the \emph{Worker Service}, which also manages liveness of the \emph{Workers} and is responsible for the distribution of tasks. There are several additional components like \emph{Front End} to visualize the parameter tuning process and \emph{Benchmarker} that automates the execution of multiple experiments and produces graphical reports out of these executions, but we leave them out of scope of this paper. 

Now let us discuss the \emph{Main Node} in more details. \emph{Experiment} is a core entity. It stores the search space and measured \emph{configurations}, parameters to be tuned and the configuration of the SPL itself.

\emph{Selection algorithm} is used to cover the search space when there is no information on its structure or the model is unreliable. Out of the box we provide two selection algorithms: Sobol sampling~\cite{sobol1999} and Fedorov's exchange algorithm~\cite{fedorov2013}, but the user can extend the SPL with their own approaches. Each selection algorithm has to provide functionality on how to get the new configuration from the search space and how to disable a point after measuring it. 

\emph{Model} is being continuously built after each measurement phase iteration, it predicts new configurations to be measured and performs self-validation. Regression model is complemented by a Bayesian optimization model inspired by~\cite{BOHB-18}.

\emph{Repeater} is a feature that automatically decides on the number of repetitions needed to obtain the required accuracy for each configuration. The basic idea is to decrease the number of repetitions for non-promising configurations, while measuring the important ones with higher preciseness. We have implemented several variants of repetition strategies. Quantity-based is a simplistic strategy, where the user specifies the number of repetitions to be performed for each configuration. Student-based repetition reduction is a strategy, that takes into account preciseness of measurements in terms of the relative measurement error. It is also possible to use knowledge available in the model to further reduce repetitions. Model-aware Student repeater does it manipulating the affordable measurement error. E.g., for those configurations which are considered not promising, the accuracy requirements are relaxed. 

\begin{figure}[h]
	\centering
	\includegraphics[width=1\linewidth]{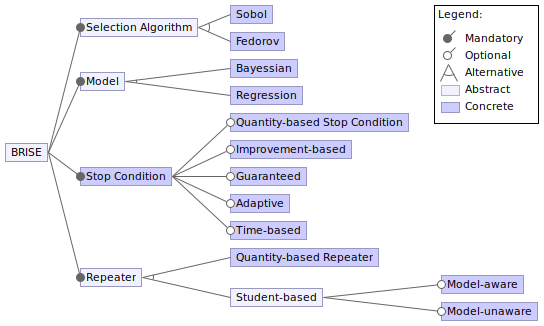}
	\caption{Feature tree of the SPL}
	\label{fig:brise-spl}
\end{figure}

\emph{Stop condition} is another mandatory feature which validates a solution received from the model and decides, whether to perform subsequent measurements or to stop the experiment. The overall Stop Condition can be combined from several stop criteria such as exceeding a specific number of measured configurations~(quantity-based), percentage of the search space~(adaptive) or time limit~(time-based) specified by the user. The lack of the result's improvement since finding the last new solution~(improvement-based) or a guaranty of finding a configuration better than the starting one~(guaranteed) are also possible stopping criteria. Figure~\ref{fig:brise-spl} summarizes the features of the SPL and their variants.

%The main drawback of using regression as the model for the search space is its weak scaling with the number of dimensions and dependence from its own hyperparameters~\cite{S-16}. Therefore, we incorporated a parameter-less approach of Bayesian optimization as described in~\cite{BOHB-18}. It uses (short on BOHB) Note, that BRISE does not restrict itself with a single model, by adding it to our framework we show its extensibility.

%% file: eval.tex
\section{Evaluation}
\label{sec:eval}

The goal of this paper is to motivate the usage of approximate solvers in combination with automatic parameter tuning based on a specific case study~(quality-based software selection and hardware mapping). Therefore, we leave a full-fledged evaluation of the SPL itself out of scope of this paper, but discuss the effect it makes on the metaheuristic-based solver and compare both variants with the ILP-based solver.

We used the same ILP solver as the authors of the approach~\cite{ttc18}, GLPK 4.65. Note, that one can use a much more efficient commercial solver which can be also tuned for performance~\cite{Hutter10} like the MH-based solver. However, as will be shown later, the main drawback of ILP-based solvers is their costly generation time. Thus, the choice of the ILP solver does not impact our findings.  

 There exist numerous recommendations for picking a default configuration of simulated annealing~\cite{Ben-Ameur2004, Kirkpatrick1987, SSB-09}, which are sometimes contradictory, e.g., in~\cite{Kirkpatrick1987} the authors suggest high starting temperature, while in~\cite{SSB-09} the authors motivate usage of smaller temperature values. We decided to take a maximum possible initial temperature, so that the algorithm starts with global exploration of the search space and resides in local regions after getting familiar with its structure. Regarding the other parameters we have decided to start with a balanced influence of picking wrong software or hardware components' on the solution's score, 1 point for an unsatisfying HWC sub-component, 5 points for an unsatisfying SWC. We have also picked the largest allowed neighborhood size and pursue the steepest descent strategy. Thus, the default~(manually tuned) configuration looks as follows:
 
\begin{itemize}
	\item \emph{subComponentUnassignedFactor}: 1;
	\item \emph{softwareComponentUnassignedFactor}: 5;
	\item \emph{hardScoreStartingTemperature}: 100;
	\item\emph{softScoreStartingTemperature}: 100;
	\item\emph{neighborhoodSize}: 50.
\end{itemize}

We configured our parameter tuning experiment in the following way. We picked Sobol selection algorithm, Bayessian model, improvement-based stop condition, which fires after 50 subsequent configurations without improvement and quantity-based repeater with 2 repetition for each configuration. For parameter tuning we decreased the timeout for running a single configuration to 10 seconds. This reduction has sped up the tuning~(the overall process of parameter tuning for all use cases took 2.34 hours), but may has influenced the result's quality as some good configurations could have been cut out by the model. After the parameter tuning stage we returned to the maximum solving time of 15 minutes as it was in the TTC case. The measurements were performed on an Intel Core i7-8700 CPU machine with 64G of memory using Fedora Server 29 with GLPK 4.65 and Oracle Java 1.8.0 201.

Table~\ref{tab:ttc} shows the results  for the scenarios introduced in~\cite{ttc18}, using the ILP-based solver~(GLPK 4.65, first value in each column), manually tuned MH-solver~(second value in each column) and automatically tuned MH-solver~(bold value). We measure the validity and quality of the result, i.e., both hard and soft score, for a corresponding problem size for each approach. Note, that solution quality is taken as a relation to the optimal solution, which was obtained by the ILP-based solver without time constrains. Moreover, we compare the time to transform the problem into a suitable format ready to be solved by the ILP and MH solvers, respectively. We also track the timestamp of the first valid solution obtained by each solver and the last improvement of the solution. Note, that the ILP solver guarantees the optimality of the obtained solution and stops after it finds it, while MH solver always runs until the timeout.

Scenarios are separated into four main groups based on the number of requests: small~(1 request), medium~(10-15 requests), large~(20 requests) and huge~(50 requests). Moreover, each group has three variations: basic, increased number of hardware resources~(\emph{much hardware}) and increased depth of the software tree~(\emph{complex software}). The resulting numbers of available implementations and total number of resources are presented in Table~\ref{tab:ttc}. To get a deeper understanding of problem scenarios, the interested reader is referred to the TTC'18 case~\cite{ttc18}.

\begin{table*}[t]
	\centering
	\caption{Comparison of ILP-based approach / manually tuned MH / automatically tuned MH}
	\label{tab:ttc}
	\resizebox{\textwidth}{!}{\begin{tabular}{rlccccccc}
		&                   &		&  Compute		&             &               & Transformation  & First valid & Last valid \\ 
		& Scenario          &	Impl's	& resources       & Valid?            & Quality              & time (s) &solution found (s) & solution found (s)  \\ 
		\midrule 
		0 & trivial                   &1&1&  \cmark / \cmark / \cmark           & 1.0 / 1.0 / \textbf{1.0}               &  0.01 / 0.01 / \textbf{0.01}        & 0.01 / 0.01 / \textbf{0.01}    &    0.01 / 0.01 / \textbf{0.01}    \\
		1 & small                     &6&5&  \cmark / \cmark / \cmark           & 1.0 / 1.0 / \textbf{1.0}                & 0.02 / 0.01 / \textbf{0.01}          & 0.02 / 4.21 / \textbf{4.21}    &   0.02 / 4.21 / \textbf{4.21}     \\
		2 & small, much hardware      &6&15&  \cmark / \cmark / \cmark           & 1.0 / 1.0 / \textbf{1.0}                &  0.03 / 0.01 / \textbf{0.01}        & 0.03 / 0.11 / \textbf{0.11}    &  0.03 / 0.11 / \textbf{0.11}      \\
		3 & small, complex software   &62&47&  \cmark / \cmark / \cmark           & 1.0 / 1.0 / \textbf{1.0}                & 0.17 / 0.01 / \textbf{0.01}        & 0.19 / 3.45 / \textbf{3.45}  &   0.19 / 890 / \textbf{890}      \\
		4 & medium                    &30&68&  \cmark / \cmark / \cmark           & 1.0 / 0.84 / \textbf{0.84}         & 1.59 / 0.01 / \textbf{0.01}        & 2.38 / 3.38 / \textbf{3.38}      & 3.3 / 881 / \textbf{881} \\
		5 & medium, much hardware     &30&225&  \cmark / \cmark / \cmark           & 1.0 / 0.94 / \textbf{0.94}         & 5.12 / 0.01 / \textbf{0.01}     & 5.72 / 1.62 / \textbf{1.62}        & 33.5 / 464 / \textbf{464} \\
		6 & medium, complex software  &155&465&  \xmark / \xmark / \xmark           & \xmark / \xmark / \xmark                & 27.8 / 0.01 / \textbf{0.01}     &  \xmark / \xmark / \xmark &    \xmark / \xmark / \xmark    \\
		7 & large                     &60&90&  \cmark / \cmark / \cmark    		  & 0.88 / 0.61 / \textbf{0.61}                & 7.75 / 0.01 / \textbf{0.01}    &  13.97 / 338 / \textbf{338} &   804 / 338 / \textbf{338}    \\
		8 & large, much hardware      &60&300&  \xmark / \cmark/ \cmark           & \xmark / 0.78 / \textbf{0.83}                & 23.55 / 0.01 / \textbf{0.01}     & \xmark / 536 / \textbf{94.5} &   \xmark / 840 / \textbf{240}    \\
		9 & large, complex software   &310&930&  \xmark / \xmark / \xmark           & \xmark / \xmark / \xmark                & \xmark / 0.02 / \textbf{0.02}   &  \xmark / \xmark / \xmark &    \xmark / \xmark / \xmark    \\
		10 & huge                     &150&225&  \xmark / \cmark / \cmark    & \xmark / 0.43 / \textbf{0.51}                & \xmark / 0.02 / \textbf{0.02}    & \xmark / 452 / \textbf{705} &   \xmark / 743 / \textbf{705}    \\
		11 & huge, much hardware      &150&750&  \xmark / \cmark / \cmark    & \xmark / 0.38 / \textbf{0.43}                & \xmark / 0.02 / \textbf{0.02}     & \xmark / 15.28 / \textbf{895} &   \xmark / 285 / \textbf{899}    \\
		12 & huge, complex software   &620&2325&  \xmark / \xmark / \xmark          & \xmark / \xmark / \xmark                & \xmark / 0.05 / \textbf{0.05}   &  \xmark / \xmark / \xmark &    \xmark / \xmark / \xmark    \\
	\end{tabular}} 
\end{table*}

From Table~\ref{tab:ttc} we can see that both ILP- and MH-based solvers are unable to solve problems \texttt{6, 9} and \texttt{12}, i.e., the problems with complex SWC structure. In terms of the MH-based solver it is determined by the move strategy of the heuristic. A move changes a single SWC at a random position in the dependency tree; thus, increasing the hard score. As a complex SWC structure results in frequent SWC-moves, the temperature should be very high; henceforth, the heuristic behaves like a random search.   

As can be seen from Table~\ref{tab:ttc}, for the smaller problem sizes \texttt{0-4} and \texttt{7} the utilization of ILP-based solver is preferential. Transformation time for these problems is very small~(less than 2 seconds), whilst solving time is even shorter. Though the MH solver has almost negligible transformation time~(<~0.1~s), it needs more time to find a valid solution and even more time for an optimal~(problem sizes \texttt{4}).

The problem size \texttt{5} is a border case, which shows the scalability behavior of both approaches. Here, the transformation time of ILP-based solver exceeds the first solution found timestamp of the MH-solver; thus, delivering the first valid solution later and is not able to deliver an optimal solution until the timeout. Although delivering a better final solution, the ILP-based solver loses to the MH-based in terms of getting the valid result. With problem size growth~(problem sizes \texttt{8, 10, 11}) we can see a development of this trend. Scenario \texttt{8} is the largest problem that can be transformed in the given time limit, but no solution is found, while for scenarios \texttt{10} and \texttt{11} the problem cannot be even transformed. On contrary, MH-based solver delivers valid solutions for all these scenarios.

\begin{figure}[t]
	\centering
	\includegraphics[width=\linewidth]{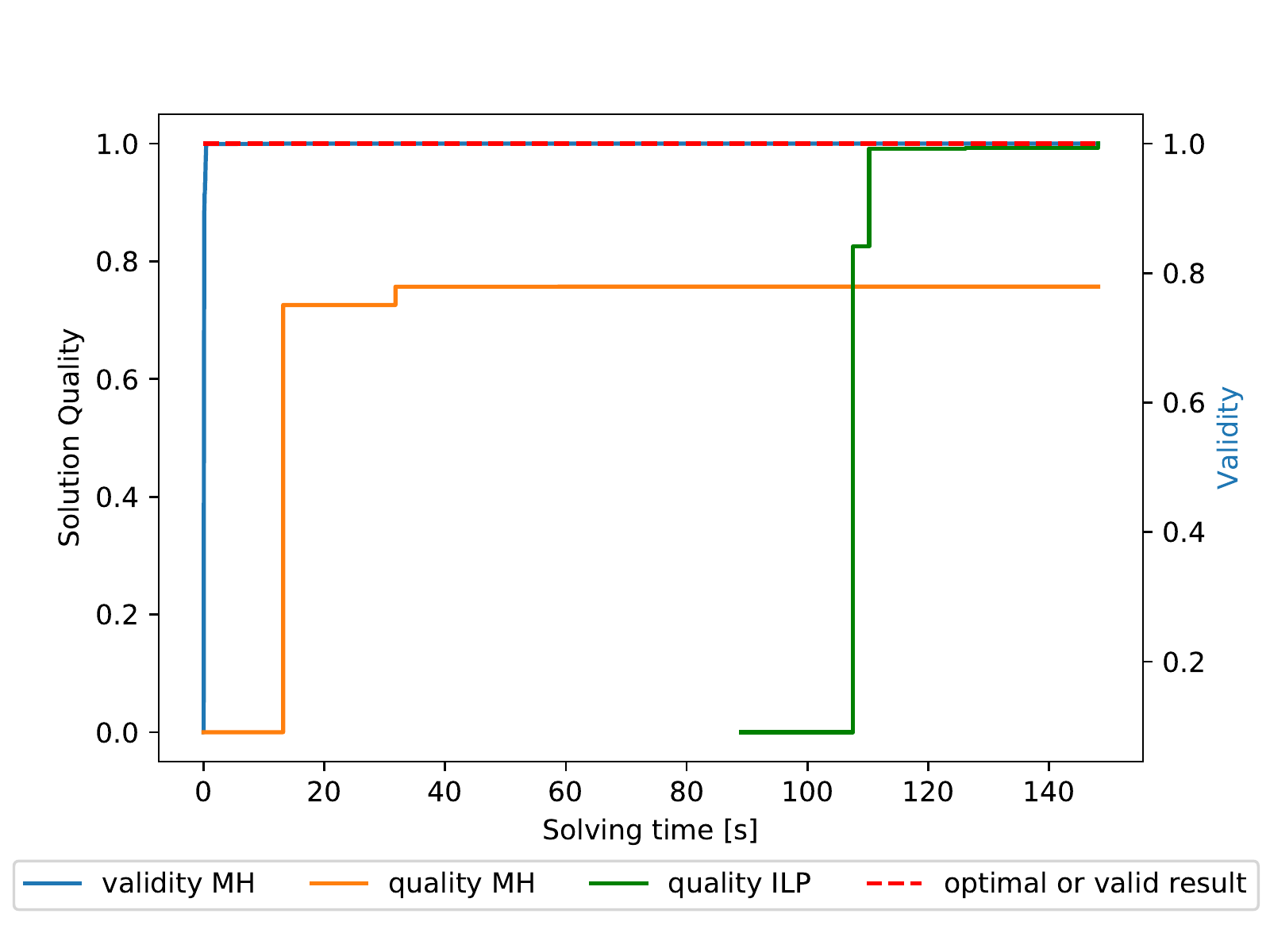}
	\caption{Solution quality improvement of MH- and ILP-based solvers}
	\label{fig:score_improvement}
\end{figure}

To illustrate this scalability trend more vividly, we stabilized the number of requests, SWCs and their structure~(4 requests each requesting a SWC chain of length 4) and scaled the number of HWCs. Figure~\ref{fig:score_improvement} shows the improvement of solutions' quality found by MH- and ILP-based solvers in comparison to an optimal result for a selected problem size~(2048 HWCs). Note, that solution's validity is normalized here, 1 means a valid solution, while 0 is the starting solution's validity. Solution's quality is taken in relation to the optimal solution. The blue line shows the change of validity of the MH solution, it instantly improves from the initial solution to an almost valid, with only several unsatisfied constraints. Being ``almost'' valid is certainly not enough and heuristic needs about 15 seconds to find the first valid solution~(see a leap of the orange line). The orange line represents the quality of the MH result, until the valid solution is found we treat the quality as zero. After the solution is found we compare its quality with an optimal result, which is about 70\%. During the optimization process heuristic finds one better solution and resides with the result of 75\% comparing to the optimum.

The ILP-based solver finds the optimal solution very fast (in about 60 seconds). However, the transformation time, which implies a construction of the ILP provides a 90 second overhead resulting in overall 150 seconds of execution. The first valid solution is already better than the one found by the heuristic, but is provided only after 110 seconds. Therefore, we can observe a window of 95 seconds, when the utilization of heuristic is beneficial as the ILP-based solver is performing costly model transformation~(\textbf{RQ2}).

\begin{figure}[t]
	\centering
	\includegraphics[width=\linewidth]{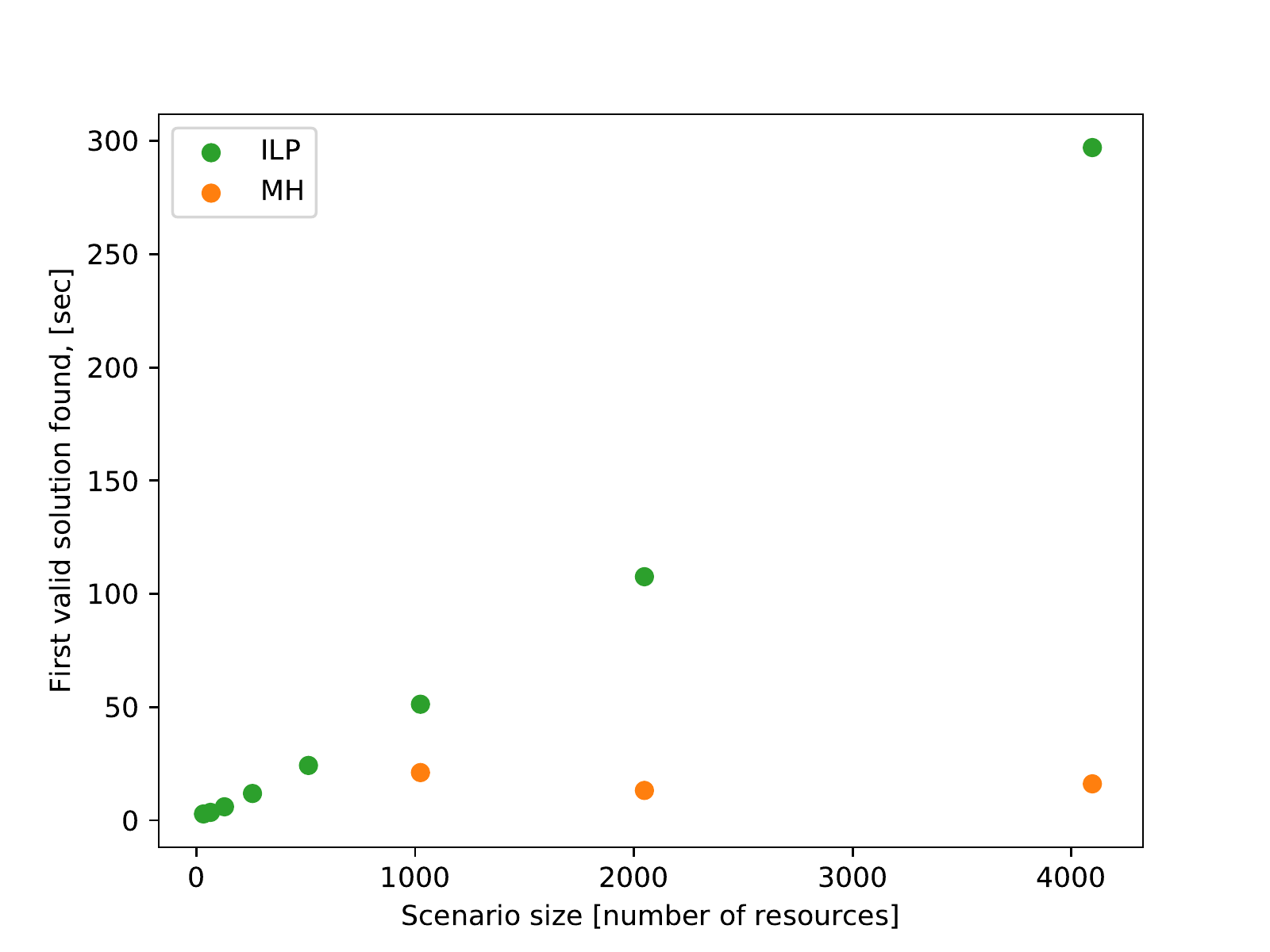}
	\caption{Execution time to find the first valid solution by MH and ILP-based solvers with changing problem size}
	\label{fig:window_scaling}
\end{figure}

Figure~\ref{fig:window_scaling} depicts the window between the first solution found by MH- and ILP-based solvers respectively and its change with the problem size. We can see that the size of the window grows exponentially with more HWC being added, thus, making the MH-based solver more and more viable~(\textbf{RQ3}). In Figure~\ref{fig:window_scaling} we can also see that the MH-based solver starts finding a solution starting from 1024 HWCs and is unable to solve smaller problems. It is a drawback of all parameter-dependent approaches, a single parameter setting cannot fit all problem sizes.

 The reason for such an inefficient scalability of the ILP-based solver's transformation time is imposed by its nature. In order to form a linear program one needs to list all the possible combinations of HWCs and SWCs in the objective function. Additionally, to form the ILP constraints one needs to evaluate each mapping. MH-based Solver, on contrary, performs the evaluation of HWC-SWC mappings only during the score calculation process, i.e., at runtime. The transformation in this case is only a single iteration through all components and thus, scales linearly.  

 Table~\ref{tab:ttc} shows us not only a comparison of MH- and ILP-based solvers, but the effect of parameter tuning and robustness of a single configuration. Manually obtained initial configuration is a good example for the robustness. From Table~\ref{tab:ttc} we can see that the manually tuned configuration can solve all the same problem instances as the automatically tuned solver. Moreover, for the smaller scenarios \texttt{1-5} and \texttt{7} BRISE 2 was not able to find a better configuration. For the scenarios \texttt{8, 10} and \texttt{11} automatic parameter tuning can provide improved solutions, 11\% better~(\textbf{RQ1}). We assume that this trend remains with scaling. The full results as well as the SPL itself are available online\footnote{\url{https://github.com/dpukhkaiev/BRISE2/tree/MoST-Rec-19}}.

%\subsection{Viability of parameter tuning}

%Last but not least, let us discuss the conditions, under which the usage of parameter tuning itself pays of.  The size of the ``viability window'' is Y minutes for \emph{large much hardware} problem size and Z minutes for \emph{large much hardware} problem. Thus, the usage of parameter tuning pays of in N iterations of optimization~(\textbf{RQ4}). However, for smaller problem sizes~(0-3 in Table~\ref{tab:ttc}) the utilization of a heuristic-based solver does not bring any benefits in comparison to ILP-based solver.

%% file: threats.tex
\section{Threats to Validity}
\label{sec:threats}

The highest threat of internal validity is an implementation of the heuristic. In general, a thorough implementation of a heuristic-based solver as well as choice of a different meta-heuristic as a basis has a major influence on the quality of the resulting solution. While developing the MH-based solver, we have developed several strategies of moves and score calculation with each subsequent iteration being more efficient. Investing more effort in solver engineering could have improved the resulting solution to a higher extent than our parameter tuning approach. However, parameter tuning can be applied on top of numerous optimization solver, thus, allowing to focus exactly on development of the solver.

Another threat is the maximal time we allowed for each configuration during tuning. We have picked 10 seconds as this cap, while in the TTC case~(production in some sense) the maximal allowed time is 15 minutes. Such a reduction could have resulted in a wrongly picked optimal configuration as some configuration may have a steadier improvement speed, but ending up in a global optimum in the end.

One more threat comes from the area of exact solvers. The main drawbacks of exact solvers are high generation time and inefficient scaling with problem size growth. Utilization of abstraction on some problem entity, e.g., packing of similar hardware components in clusters may decrease both generation and solving time, while decreasing solution quality to some extent.

A threat to external validity lies in utilizing a single use case as a case study in this paper. We have still yet to define the limits of successful application of parameter tuning for combinatorial optimization. 

%% file: conclusion.tex
\section{Conclusion and Future Work}
\label{sec:conclusion}

Utilization of heuristic-based approaches for combinatorial optimization problems has its benefits and drawbacks. Active learning approaches to parameter tuning can aid the researcher to focus on improving the solver itself, while leaving a costly and often not obvious choice of parameters to the tuner. In this paper we present a simple metaheuristic-based solver to quality-based software selection and hardware mapping problem that enhanced with a parameter tuning software product line is able to outperform the state-of-the-art exact approaches in terms of scaling and time of finding the first valid solution, while the same solver being 11\% less efficient without tuning.

Our future work goes in several directions. We will continue extending the software product line with new features, such as, outlier detection, multi-objectiveness and adaptive task interruption strategies. We will try to push the limits of parameter tuning to combinatorial optimization problems themselves by decomposing variant selection and resource allocation into separate stages, thus, trading of exploration~(variant selection) and exploitation~(resource allocation). Exploration is a natural task of the \emph{model}, while exploitation can be treated as a separate optimization problem.

Another promising research direction lies in abstraction of a problem formulation of an exact solver. Here, we aim to increase the performance of the approaches, whilst trading-off optimality.

%% file: acknowledgements.tex
\section*{Acknowledgments}
This work is supported (in part) by the German Research Foundation (DFG) within the Collaborative Research Center SFB 912--HAEC and  BMBF project "Software Campus" (grant no. 01IS17044).

%% file: mquat-mds.bbl

\begin{thebibliography}{32}

%%% the Web address.

\ifx \showCODEN    \undefined \def \showCODEN     #1{\unskip}     \fi
\ifx \showDOI      \undefined \def \showDOI       #1{#1}\fi
\ifx \showISBNx    \undefined \def \showISBNx     #1{\unskip}     \fi
\ifx \showISBNxiii \undefined \def \showISBNxiii  #1{\unskip}     \fi
\ifx \showISSN     \undefined \def \showISSN      #1{\unskip}     \fi
\ifx \showLCCN     \undefined \def \showLCCN      #1{\unskip}     \fi
\ifx \shownote     \undefined \def \shownote      #1{#1}          \fi
\ifx \showarticletitle \undefined \def \showarticletitle #1{#1}   \fi
\ifx \showURL      \undefined \def \showURL       {\relax}        \fi
% invisible to TeX
\providecommand\bibfield[2]{#2}
\providecommand\bibinfo[2]{#2}
\providecommand\natexlab[1]{#1}
\providecommand\showeprint[2][]{arXiv:#2}

\bibitem[\protect\citeauthoryear{Alipourfard, Liu, Chen, Venkataraman, Yu, and
  Zhang}{Alipourfard et~al\mbox{.}}{2017}]%
        {Alipourfard2017}
\bibfield{author}{\bibinfo{person}{Omid Alipourfard},
  \bibinfo{person}{Hongqiang~Harry Liu}, \bibinfo{person}{Jianshu Chen},
  \bibinfo{person}{Shivaram Venkataraman}, \bibinfo{person}{Minlan Yu}, {and}
  \bibinfo{person}{Ming Zhang}.} \bibinfo{year}{2017}\natexlab{}.
\newblock \showarticletitle{Cherrypick: Adaptively Unearthing the Best Cloud
  Configurations for Big Data Analytics}. In \bibinfo{booktitle}{{\em
  Proceedings of the 14th USENIX Conference on Networked Systems Design and
  Implementation}} {\em (\bibinfo{series}{NSDI'17})}.
  \bibinfo{publisher}{USENIX Association}, \bibinfo{address}{Berkeley, CA,
  USA}, \bibinfo{pages}{469--482}.
\newblock
\showISBNx{978-1-931971-37-9}
\showURL{%
\url{http://dl.acm.org/citation.cfm?id=3154630.3154669}}


\bibitem[\protect\citeauthoryear{Antosiewicz, Koloch, and
  Kamiński}{Antosiewicz et~al\mbox{.}}{2013}]%
        {Antosiewicz2013}
\bibfield{author}{\bibinfo{person}{M. Antosiewicz}, \bibinfo{person}{G.
  Koloch}, {and} \bibinfo{person}{B. Kamiński}.}
  \bibinfo{year}{2013}\natexlab{}.
\newblock \showarticletitle{Choice of best possible metaheuristic algorithm for
  the travelling salesman problem with limited computational time: quality,
  uncertainty and speed}.
\newblock \bibinfo{journal}{{\em Journal of Theoretical and Applied Computer
  Science\/}}  \bibinfo{volume}{Vol. 7, nr 1} (\bibinfo{year}{2013}),
  \bibinfo{pages}{46--55}.
\newblock


\bibitem[\protect\citeauthoryear{Arcuri and Fraser}{Arcuri and Fraser}{2011}]%
        {AFC11}
\bibfield{author}{\bibinfo{person}{Andrea Arcuri} {and} \bibinfo{person}{Gordon
  Fraser}.} \bibinfo{year}{2011}\natexlab{}.
\newblock \showarticletitle{On Parameter Tuning in Search Based Software
  Engineering}. In \bibinfo{booktitle}{{\em Search Based Software
  Engineering}}, \bibfield{editor}{\bibinfo{person}{Myra~B. Cohen} {and}
  \bibinfo{person}{Mel {\'O}~Cinn{\'e}ide}} (Eds.).
  \bibinfo{publisher}{Springer Berlin Heidelberg}, \bibinfo{address}{Berlin,
  Heidelberg}, \bibinfo{pages}{33--47}.
\newblock
\showISBNx{978-3-642-23716-4}


\bibitem[\protect\citeauthoryear{Ben-Ameur}{Ben-Ameur}{2004}]%
        {Ben-Ameur2004}
\bibfield{author}{\bibinfo{person}{Walid Ben-Ameur}.}
  \bibinfo{year}{2004}\natexlab{}.
\newblock \showarticletitle{Computing the Initial Temperature of Simulated
  Annealing}.
\newblock \bibinfo{journal}{{\em Computational Optimization and
  Applications\/}} \bibinfo{volume}{29}, \bibinfo{number}{3} (\bibinfo{date}{01
  Dec} \bibinfo{year}{2004}), \bibinfo{pages}{369--385}.
\newblock
\showISSN{1573-2894}
\showDOI{%
\url{https://doi.org/10.1023/B:COAP.0000044187.23143.bd}}


\bibitem[\protect\citeauthoryear{Birattari}{Birattari}{2009}]%
        {Birattari2009}
\bibfield{author}{\bibinfo{person}{Mauro Birattari}.}
  \bibinfo{year}{2009}\natexlab{}.
\newblock \bibinfo{booktitle}{{\em Tuning Metaheuristics: A Machine Learning
  Perspective\/} (\bibinfo{edition}{1st ed. 2005. 2nd printing} ed.)}.
\newblock \bibinfo{publisher}{Springer Publishing Company, Incorporated}.
\newblock
\showISBNx{3642004822, 9783642004827}


\bibitem[\protect\citeauthoryear{Boussaïd, Lepagnot, and Siarry}{Boussaïd
  et~al\mbox{.}}{2013}]%
        {BOUSSAID201382}
\bibfield{author}{\bibinfo{person}{Ilhem Boussaïd}, \bibinfo{person}{Julien
  Lepagnot}, {and} \bibinfo{person}{Patrick Siarry}.}
  \bibinfo{year}{2013}\natexlab{}.
\newblock \showarticletitle{A survey on optimization metaheuristics}.
\newblock \bibinfo{journal}{{\em Information Sciences\/}}
  \bibinfo{volume}{237} (\bibinfo{year}{2013}), \bibinfo{pages}{82 -- 117}.
\newblock
\showISSN{0020-0255}
\showDOI{%
\url{https://doi.org/10.1016/j.ins.2013.02.041}}


\bibitem[\protect\citeauthoryear{Camilleri, Neri, and Papoutsidakis}{Camilleri
  et~al\mbox{.}}{2014}]%
        {camilleri2014}
\bibfield{author}{\bibinfo{person}{Michel Camilleri}, \bibinfo{person}{Filippo
  Neri}, {and} \bibinfo{person}{MICHALIS Papoutsidakis}.}
  \bibinfo{year}{2014}\natexlab{}.
\newblock \showarticletitle{An algorithmic approach to parameter selection in
  machine learning using meta-optimization techniques}.
\newblock \bibinfo{journal}{{\em WSEAS Transactions on systems\/}}
  \bibinfo{volume}{13} (\bibinfo{year}{2014}), \bibinfo{pages}{202 -- 213}.
\newblock


\bibitem[\protect\citeauthoryear{Dalibard, Schaarschmidt, and Yoneki}{Dalibard
  et~al\mbox{.}}{2017}]%
        {Dalibard2017}
\bibfield{author}{\bibinfo{person}{Valentin Dalibard}, \bibinfo{person}{Michael
  Schaarschmidt}, {and} \bibinfo{person}{Eiko Yoneki}.}
  \bibinfo{year}{2017}\natexlab{}.
\newblock \showarticletitle{BOAT: Building Auto-Tuners with Structured Bayesian
  Optimization}. In \bibinfo{booktitle}{{\em Proceedings of the 26th
  International Conference on World Wide Web}} {\em (\bibinfo{series}{WWW
  '17})}. \bibinfo{publisher}{International World Wide Web Conferences Steering
  Committee}, \bibinfo{address}{Republic and Canton of Geneva, Switzerland},
  \bibinfo{pages}{479--488}.
\newblock
\showISBNx{978-1-4503-4913-0}
\showDOI{%
\url{https://doi.org/10.1145/3038912.3052662}}


\bibitem[\protect\citeauthoryear{Dean, Morris, Stufken, and Bingham}{Dean
  et~al\mbox{.}}{2015}]%
        {dean2015handbook}
\bibfield{author}{\bibinfo{person}{Angela Dean}, \bibinfo{person}{Max Morris},
  \bibinfo{person}{John Stufken}, {and} \bibinfo{person}{Derek Bingham}.}
  \bibinfo{year}{2015}\natexlab{}.
\newblock \bibinfo{booktitle}{{\em Handbook of design and analysis of
  experiments}}. Vol.~\bibinfo{volume}{7}.
\newblock \bibinfo{publisher}{CRC Press}.
\newblock


\bibitem[\protect\citeauthoryear{Falkner, Klein, and Hutter}{Falkner
  et~al\mbox{.}}{2018}]%
        {BOHB-18}
\bibfield{author}{\bibinfo{person}{Stefan Falkner}, \bibinfo{person}{Aaron
  Klein}, {and} \bibinfo{person}{Frank Hutter}.}
  \bibinfo{year}{2018}\natexlab{}.
\newblock \showarticletitle{{BOHB}: Robust and Efficient Hyperparameter
  Optimization at Scale}. In \bibinfo{booktitle}{{\em Proceedings of the 35th
  International Conference on Machine Learning}} {\em
  (\bibinfo{series}{Proceedings of Machine Learning Research})},
  \bibfield{editor}{\bibinfo{person}{Jennifer Dy} {and}
  \bibinfo{person}{Andreas Krause}} (Eds.), Vol.~\bibinfo{volume}{80}.
  \bibinfo{publisher}{PMLR}, \bibinfo{address}{Stockholmsmässan, Stockholm
  Sweden}, \bibinfo{pages}{1437--1446}.
\newblock
\showURL{%
\url{http://proceedings.mlr.press/v80/falkner18a.html}}


\bibitem[\protect\citeauthoryear{Fedorov}{Fedorov}{2013}]%
        {fedorov2013}
\bibfield{author}{\bibinfo{person}{Valerii~Vadimovich Fedorov}.}
  \bibinfo{year}{2013}\natexlab{}.
\newblock \bibinfo{booktitle}{{\em Theory of optimal experiments}}.
\newblock \bibinfo{publisher}{Elsevier}.
\newblock


\bibitem[\protect\citeauthoryear{Feurer and Hutter}{Feurer and Hutter}{2019}]%
        {Feurer2019}
\bibfield{author}{\bibinfo{person}{Matthias Feurer} {and}
  \bibinfo{person}{Frank Hutter}.} \bibinfo{year}{2019}\natexlab{}.
\newblock \bibinfo{booktitle}{{\em Hyperparameter Optimization}}.
\newblock \bibinfo{publisher}{Springer International Publishing},
  \bibinfo{address}{Cham}, \bibinfo{pages}{3--33}.
\newblock
\showISBNx{978-3-030-05318-5}
\showDOI{%
\url{https://doi.org/10.1007/978-3-030-05318-5_1}}


\bibitem[\protect\citeauthoryear{Golovin, Solnik, Moitra, Kochanski, Karro, and
  Sculley}{Golovin et~al\mbox{.}}{2017}]%
        {Golovin2017}
\bibfield{author}{\bibinfo{person}{Daniel Golovin}, \bibinfo{person}{Benjamin
  Solnik}, \bibinfo{person}{Subhodeep Moitra}, \bibinfo{person}{Greg
  Kochanski}, \bibinfo{person}{John Karro}, {and} \bibinfo{person}{D.
  Sculley}.} \bibinfo{year}{2017}\natexlab{}.
\newblock \showarticletitle{Google Vizier: A Service for Black-Box
  Optimization}. In \bibinfo{booktitle}{{\em Proceedings of the 23rd ACM SIGKDD
  International Conference on Knowledge Discovery and Data Mining}} {\em
  (\bibinfo{series}{KDD '17})}. \bibinfo{publisher}{ACM}, \bibinfo{address}{New
  York, NY, USA}, \bibinfo{pages}{1487--1495}.
\newblock
\showISBNx{978-1-4503-4887-4}
\showDOI{%
\url{https://doi.org/10.1145/3097983.3098043}}


\bibitem[\protect\citeauthoryear{G{\"o}tz, K{\"u}hn, Piechnick, P{\"u}schel,
  and A{\ss}mann}{G{\"o}tz et~al\mbox{.}}{2014}]%
        {GKPPA-14}
\bibfield{author}{\bibinfo{person}{Sebastian G{\"o}tz}, \bibinfo{person}{Thomas
  K{\"u}hn}, \bibinfo{person}{Christian Piechnick}, \bibinfo{person}{Georg
  P{\"u}schel}, {and} \bibinfo{person}{Uwe A{\ss}mann}.}
  \bibinfo{year}{2014}\natexlab{}.
\newblock \showarticletitle{A Models@run.time Approach for Multi-objective
  Self-optimizing Software}. In \bibinfo{booktitle}{{\em Adaptive and
  Intelligent Systems}}, \bibfield{editor}{\bibinfo{person}{Abdelhamid
  Bouchachia}} (Ed.). \bibinfo{publisher}{Springer International Publishing},
  \bibinfo{address}{Cham}, \bibinfo{pages}{100--109}.
\newblock
\showISBNx{978-3-319-11298-5}


\bibitem[\protect\citeauthoryear{G{\"o}tz, Mey, Sch{\"o}ne, and
  A{\ss}mann}{G{\"o}tz et~al\mbox{.}}{2018}]%
        {ttc18}
\bibfield{author}{\bibinfo{person}{Sebastian G{\"o}tz},
  \bibinfo{person}{Johannes Mey}, \bibinfo{person}{Rene Sch{\"o}ne}, {and}
  \bibinfo{person}{Uwe A{\ss}mann}.} \bibinfo{year}{2018}\natexlab{}.
\newblock \showarticletitle{Quality-based Software-Selection and
  Hardware-Mapping as Model Transformation Problem}.
\newblock In \bibinfo{booktitle}{{\em Proceedings of the Transformation Tool
  Contest, a part of the Software Technologies: Applications and Foundations
  (STAF)}}, \bibfield{editor}{\bibinfo{person}{Antonio Garcia-Dominguez},
  \bibinfo{person}{Georg Hinkel}, {and} \bibinfo{person}{Filip K\v{r}ikava}}
  (Eds.). \bibinfo{publisher}{CEUR-WS.org}.
\newblock


\bibitem[\protect\citeauthoryear{G\"{o}tz, Mey, Schöne, and Aßmann}{G\"{o}tz
  et~al\mbox{.}}{2018}]%
        {GMSA-18}
\bibfield{author}{\bibinfo{person}{Sebastian G\"{o}tz},
  \bibinfo{person}{Johannes Mey}, \bibinfo{person}{René Schöne}, {and}
  \bibinfo{person}{Uwe Aßmann}.} \bibinfo{year}{2018}\natexlab{}.
\newblock \showarticletitle{A {JastAdd}- and {ILP}-Based Solution to the
  Software-Selection and Hardware-Mapping-Problem at the {TTC} 2018}.
\newblock In \bibinfo{booktitle}{{\em Proceedings of the Transformation Tool
  Contest, a part of the Software Technologies: Applications and Foundations
  (STAF)}}, \bibfield{editor}{\bibinfo{person}{Antonio Garcia-Dominguez},
  \bibinfo{person}{Georg Hinkel}, {and} \bibinfo{person}{Filip K\v{r}ikava}}
  (Eds.). \bibinfo{publisher}{CEUR-WS.org}.
\newblock


\bibitem[\protect\citeauthoryear{HoseinDoost, Karimi, Kolahdouz-Rahimi, and
  Zamani}{HoseinDoost et~al\mbox{.}}{2018}]%
        {HD18}
\bibfield{author}{\bibinfo{person}{Samaneh HoseinDoost},
  \bibinfo{person}{Meysam Karimi}, \bibinfo{person}{Shekoufeh
  Kolahdouz-Rahimi}, {and} \bibinfo{person}{Bahman Zamani}.}
  \bibinfo{year}{2018}\natexlab{}.
\newblock \showarticletitle{Solving the Quality-based Software-Selection and
  Hardware-Mapping Problem with ACO}.
\newblock In \bibinfo{booktitle}{{\em Proceedings of the Transformation Tool
  Contest, a part of the Software Technologies: Applications and Foundations
  (STAF)}}, \bibfield{editor}{\bibinfo{person}{Antonio Garcia-Dominguez},
  \bibinfo{person}{Georg Hinkel}, {and} \bibinfo{person}{Filip K\v{r}ikava}}
  (Eds.). \bibinfo{publisher}{CEUR-WS.org}.
\newblock


\bibitem[\protect\citeauthoryear{Huang and Chang}{Huang and Chang}{2007}]%
        {HUANG2007516}
\bibfield{author}{\bibinfo{person}{Hui-Ling Huang} {and}
  \bibinfo{person}{Fang-Lin Chang}.} \bibinfo{year}{2007}\natexlab{}.
\newblock \showarticletitle{ESVM: Evolutionary support vector machine for
  automatic feature selection and classification of microarray data}.
\newblock \bibinfo{journal}{{\em Biosystems\/}} \bibinfo{volume}{90},
  \bibinfo{number}{2} (\bibinfo{year}{2007}), \bibinfo{pages}{516 -- 528}.
\newblock
\showISSN{0303-2647}
\showDOI{%
\url{https://doi.org/10.1016/j.biosystems.2006.12.003}}


\bibitem[\protect\citeauthoryear{Hutter, Hoos, and Leyton-Brown}{Hutter
  et~al\mbox{.}}{2010}]%
        {Hutter10}
\bibfield{author}{\bibinfo{person}{Frank Hutter}, \bibinfo{person}{Holger~H.
  Hoos}, {and} \bibinfo{person}{Kevin Leyton-Brown}.}
  \bibinfo{year}{2010}\natexlab{}.
\newblock \showarticletitle{Automated Configuration of Mixed Integer
  Programming Solvers}. In \bibinfo{booktitle}{{\em Integration of AI and OR
  Techniques in Constraint Programming for Combinatorial Optimization
  Problems}}, \bibfield{editor}{\bibinfo{person}{Andrea Lodi},
  \bibinfo{person}{Michela Milano}, {and} \bibinfo{person}{Paolo Toth}} (Eds.).
  \bibinfo{publisher}{Springer Berlin Heidelberg}, \bibinfo{address}{Berlin,
  Heidelberg}, \bibinfo{pages}{186--202}.
\newblock
\showISBNx{978-3-642-13520-0}


\bibitem[\protect\citeauthoryear{Jugovac, Jannach, and Lerche}{Jugovac
  et~al\mbox{.}}{2017}]%
        {JJL17}
\bibfield{author}{\bibinfo{person}{Michael Jugovac}, \bibinfo{person}{Dietmar
  Jannach}, {and} \bibinfo{person}{Lukas Lerche}.}
  \bibinfo{year}{2017}\natexlab{}.
\newblock \showarticletitle{Efficient optimization of multiple recommendation
  quality factors according to individual user tendencies}.
\newblock \bibinfo{journal}{{\em Expert Systems with Applications\/}}
  \bibinfo{volume}{81} (\bibinfo{year}{2017}), \bibinfo{pages}{321 -- 331}.
\newblock
\showISSN{0957-4174}
\showDOI{%
\url{https://doi.org/10.1016/j.eswa.2017.03.055}}


\bibitem[\protect\citeauthoryear{Kendall, Knust, Ribeiro, and Urrutia}{Kendall
  et~al\mbox{.}}{2010}]%
        {KENDALL20101}
\bibfield{author}{\bibinfo{person}{Graham Kendall}, \bibinfo{person}{Sigrid
  Knust}, \bibinfo{person}{Celso~C. Ribeiro}, {and} \bibinfo{person}{Sebastián
  Urrutia}.} \bibinfo{year}{2010}\natexlab{}.
\newblock \showarticletitle{Scheduling in sports: An annotated bibliography}.
\newblock \bibinfo{journal}{{\em Computers and Operations Research\/}}
  \bibinfo{volume}{37}, \bibinfo{number}{1} (\bibinfo{year}{2010}),
  \bibinfo{pages}{1 -- 19}.
\newblock
\showISSN{0305-0548}
\showDOI{%
\url{https://doi.org/10.1016/j.cor.2009.05.013}}


\bibitem[\protect\citeauthoryear{Khachaturyan, Semenovskaya, and
  Vainstein}{Khachaturyan et~al\mbox{.}}{1979}]%
        {khachaturyan1979statistical}
\bibfield{author}{\bibinfo{person}{A Khachaturyan}, \bibinfo{person}{S
  Semenovskaya}, {and} \bibinfo{person}{B Vainstein}.}
  \bibinfo{year}{1979}\natexlab{}.
\newblock \showarticletitle{Statistical-thermodynamic approach to determination
  of structure amplitude phases}.
\newblock \bibinfo{journal}{{\em Sov. Phys. Crystallography\/}}
  \bibinfo{volume}{24}, \bibinfo{number}{5} (\bibinfo{year}{1979}),
  \bibinfo{pages}{519--524}.
\newblock


\bibitem[\protect\citeauthoryear{Kirkpatrick, Gelatt, and Vecchi}{Kirkpatrick
  et~al\mbox{.}}{1987}]%
        {Kirkpatrick1987}
\bibfield{author}{\bibinfo{person}{S. Kirkpatrick}, \bibinfo{person}{C.~D.
  Gelatt, Jr.}, {and} \bibinfo{person}{M.~P. Vecchi}.}
  \bibinfo{year}{1987}\natexlab{}.
\newblock \showarticletitle{Readings in Computer Vision: Issues, Problems,
  Principles, and Paradigms}.
\newblock \bibinfo{publisher}{Morgan Kaufmann Publishers Inc.},
  \bibinfo{address}{San Francisco, CA, USA}, Chapter Optimization by Simulated
  Annealing, \bibinfo{pages}{606--615}.
\newblock
\showISBNx{0-934613-33-8}
\showURL{%
\url{http://dl.acm.org/citation.cfm?id=33517.33566}}


\bibitem[\protect\citeauthoryear{Kuenne and Soland}{Kuenne and Soland}{1972}]%
        {Kuenne1972}
\bibfield{author}{\bibinfo{person}{Robert~E. Kuenne} {and}
  \bibinfo{person}{Richard~M. Soland}.} \bibinfo{year}{1972}\natexlab{}.
\newblock \showarticletitle{Exact and approximate solutions to the multisource
  weber problem}.
\newblock \bibinfo{journal}{{\em Mathematical Programming\/}}
  \bibinfo{volume}{3}, \bibinfo{number}{1} (\bibinfo{date}{01 Dec}
  \bibinfo{year}{1972}), \bibinfo{pages}{193--209}.
\newblock
\showISSN{1436-4646}
\showDOI{%
\url{https://doi.org/10.1007/BF01584989}}


\bibitem[\protect\citeauthoryear{Pinedo}{Pinedo}{2012}]%
        {pinedo2012scheduling}
\bibfield{author}{\bibinfo{person}{Michael Pinedo}.}
  \bibinfo{year}{2012}\natexlab{}.
\newblock \bibinfo{booktitle}{{\em Scheduling}}.
\newblock \bibinfo{publisher}{Springer}.
\newblock


\bibitem[\protect\citeauthoryear{Pukhkaiev and G\"{o}tz}{Pukhkaiev and
  G\"{o}tz}{2018}]%
        {PG18}
\bibfield{author}{\bibinfo{person}{Dmytro Pukhkaiev} {and}
  \bibinfo{person}{Sebastian G\"{o}tz}.} \bibinfo{year}{2018}\natexlab{}.
\newblock \showarticletitle{{BRISE}: Energy-efficient Benchmark Reduction}. In
  \bibinfo{booktitle}{{\em Proceedings of the 6th International Workshop on
  Green and Sustainable Software}} {\em (\bibinfo{series}{GREENS '18})}.
  \bibinfo{publisher}{ACM}, \bibinfo{address}{New York, NY, USA},
  \bibinfo{pages}{23--30}.
\newblock
\showISBNx{978-1-4503-5732-6}
\showDOI{%
\url{https://doi.org/10.1145/3194078.3194082}}


\bibitem[\protect\citeauthoryear{{Shakouri G.}, {Shojaee}, and {Behnam
  T.}}{{Shakouri G.} et~al\mbox{.}}{2009}]%
        {SSB-09}
\bibfield{author}{\bibinfo{person}{H. {Shakouri G.}}, \bibinfo{person}{K.
  {Shojaee}}, {and} \bibinfo{person}{M. {Behnam T.}}}
  \bibinfo{year}{2009}\natexlab{}.
\newblock \showarticletitle{Investigation on the choice of the initial
  temperature in the Simulated Annealing: A mushy state SA for TSP}. In
  \bibinfo{booktitle}{{\em 2009 17th Mediterranean Conference on Control and
  Automation}}. \bibinfo{pages}{1050--1055}.
\newblock
\showDOI{%
\url{https://doi.org/10.1109/MED.2009.5164685}}


\bibitem[\protect\citeauthoryear{Sobol and Levitan}{Sobol and Levitan}{1999}]%
        {sobol1999}
\bibfield{author}{\bibinfo{person}{IM Sobol} {and} \bibinfo{person}{Yu~L
  Levitan}.} \bibinfo{year}{1999}\natexlab{}.
\newblock \showarticletitle{A pseudo-random number generator for personal
  computers}.
\newblock \bibinfo{journal}{{\em Computers \& Mathematics with Applications\/}}
  \bibinfo{volume}{37}, \bibinfo{number}{4-5} (\bibinfo{year}{1999}),
  \bibinfo{pages}{33--40}.
\newblock


\bibitem[\protect\citeauthoryear{Tang, Dai, Sun, and Meng}{Tang
  et~al\mbox{.}}{2018}]%
        {TANG201847}
\bibfield{author}{\bibinfo{person}{Xiaochuan Tang}, \bibinfo{person}{Yuanshun
  Dai}, \bibinfo{person}{Peng Sun}, {and} \bibinfo{person}{Sa Meng}.}
  \bibinfo{year}{2018}\natexlab{}.
\newblock \showarticletitle{Interaction-based feature selection using Factorial
  Design}.
\newblock \bibinfo{journal}{{\em Neurocomputing\/}}  \bibinfo{volume}{281}
  (\bibinfo{year}{2018}), \bibinfo{pages}{47 -- 54}.
\newblock
\showISSN{0925-2312}
\showDOI{%
\url{https://doi.org/10.1016/j.neucom.2017.11.058}}


\bibitem[\protect\citeauthoryear{Wan, Vahidi, and Luckow}{Wan
  et~al\mbox{.}}{2016}]%
        {WAN2016548}
\bibfield{author}{\bibinfo{person}{Nianfeng Wan}, \bibinfo{person}{Ardalan
  Vahidi}, {and} \bibinfo{person}{Andre Luckow}.}
  \bibinfo{year}{2016}\natexlab{}.
\newblock \showarticletitle{Optimal speed advisory for connected vehicles in
  arterial roads and the impact on mixed traffic}.
\newblock \bibinfo{journal}{{\em Transportation Research Part C: Emerging
  Technologies\/}}  \bibinfo{volume}{69} (\bibinfo{year}{2016}),
  \bibinfo{pages}{548 -- 563}.
\newblock
\showISSN{0968-090X}
\showDOI{%
\url{https://doi.org/10.1016/j.trc.2016.01.011}}


\bibitem[\protect\citeauthoryear{Wortmann, Waibel, Nannicini, Evins,
  Schroepfer, and Carmeliet}{Wortmann et~al\mbox{.}}{2017}]%
        {Wortmann2017}
\bibfield{author}{\bibinfo{person}{Thomas Wortmann}, \bibinfo{person}{Christoph
  Waibel}, \bibinfo{person}{Giacomo Nannicini}, \bibinfo{person}{Ralph Evins},
  \bibinfo{person}{Thomas Schroepfer}, {and} \bibinfo{person}{Jan Carmeliet}.}
  \bibinfo{year}{2017}\natexlab{}.
\newblock \showarticletitle{Are Genetic Algorithms Really the Best Choice for
  Building Energy Optimization?}. In \bibinfo{booktitle}{{\em Proceedings of
  the Symposium on Simulation for Architecture and Urban Design}} {\em
  (\bibinfo{series}{SIMAUD '17})}. \bibinfo{publisher}{Society for Computer
  Simulation International}, \bibinfo{address}{San Diego, CA, USA}, Article
  \bibinfo{articleno}{6}, \bibinfo{numpages}{8}~pages.
\newblock
\showISBNx{978-1-5108-7018-5}
\showURL{%
\url{http://dl.acm.org/citation.cfm?id=3289787.3289793}}


\bibitem[\protect\citeauthoryear{Yi, Li, Tang, and Chen}{Yi
  et~al\mbox{.}}{2015}]%
        {YI2015256}
\bibfield{author}{\bibinfo{person}{Qian Yi}, \bibinfo{person}{Congbo Li},
  \bibinfo{person}{Ying Tang}, {and} \bibinfo{person}{Xingzheng Chen}.}
  \bibinfo{year}{2015}\natexlab{}.
\newblock \showarticletitle{Multi-objective parameter optimization of CNC
  machining for low carbon manufacturing}.
\newblock \bibinfo{journal}{{\em Journal of Cleaner Production\/}}
  \bibinfo{volume}{95} (\bibinfo{year}{2015}), \bibinfo{pages}{256 -- 264}.
\newblock
\showISSN{0959-6526}
\showDOI{%
\url{https://doi.org/10.1016/j.jclepro.2015.02.076}}


\end{thebibliography}
